\documentclass[11pt,a4paper]{article}
\usepackage[hyperref]{acl2021}
\usepackage{times}
\usepackage{latexsym}
\usepackage{makecell}
\usepackage{xcolor,colortbl}
\usepackage{multirow}
\usepackage{tikz}
\usepackage{amssymb}
\usepackage{pgfplots}
\pgfplotsset{width=7.5cm,compat=1.12}
\usepgfplotslibrary{fillbetween}

\usepackage{amsmath}

\usepackage{hhline}
\usepackage{microtype}
\newcolumntype{M}[1]{>{\centering\arraybackslash}m{#1}}
\definecolor{mygreen}{RGB}{185,224,165}
\definecolor{myred}{RGB}{241,156,153}

\aclfinalcopy 

\title{W-RST: Towards a Weighted RST-style Discourse Framework}

\author{Patrick Huber\thanks{~~Equal contribution.}, Wen Xiao\footnotemark[1], Giuseppe Carenini\\
  Department of Computer Science \\
  University of British Columbia \\
  Vancouver, BC, Canada, V6T 1Z4 \\
  {\tt \{huberpat, xiaowen3, carenini\}@cs.ubc.ca}}

\date{}

\begin{document}
\maketitle
\begin{abstract}

Aiming for a better integration of data-driven and linguistically-inspired approaches,
we explore whether 
RST Nuclearity, 
assigning a binary assessment of importance between text segments, 
can be replaced by automatically generated, real-valued 
scores, in what we call a Weighted-RST 
framework. 
In particular, we find that weighted discourse trees 
from auxiliary tasks can benefit key NLP downstream applications compared to nuclearity-centered approaches. We further show that real-valued importance distributions partially and interestingly align with the assessment and uncertainty of human annotators.

\end{abstract}

\section{Introduction}

Ideally, research in Natural Language Processing (NLP) 
should balance and integrate findings from 
machine learning approaches 
with insights and theories from linguistics. 
With the enormous success of data-driven approaches over the last decades, this balance has arguably and excessively 
shifted, with linguistic theories playing a less and less critical role. Even more importantly, there are only little attempts made to improve such theories in light of recent empirical results.

In the context of discourse, two main theories have emerged in the past: The Rhetorical Structure Theory (RST) \cite{carlson2002rst} and PDTB \cite{prasadpenn}. In this paper, we focus on RST, exploring whether the underlying theory can be refined in a data-driven manner.

In general, RST postulates a complete discourse tree for a given document. To obtain this formal representation 
as a projective consituency tree, 
a given document is first separated into so called Elementary Discourse Units (or short EDUs), representing clause-like sentence fragments of the input document. Afterwards, the discourse tree is built by hierarchically aggregating EDUs into larger constituents annotated with an importance indicator (in RST called nuclearity) and a relation holding between siblings in the aggregation. The nuclearity attribute in RST thereby assigns each sub-tree either a nucleus-attribute, indicating central importance of the sub-tree in the context of the document, or a satellite-attribute, categorizing the sub-tree as of peripheral importance. 
The relation attribute further 
characterizes the connection between sub-trees (e.g. Elaboration, Cause, Contradiction).

One central requirement of the RST discourse theory, as for all linguistic theories, is that a trained human should be able to specify and interpret the discourse representations. While this is a clear advantage when trying to generate explainable outcomes, it also introduces problematic, 
human-centered simplifications; 
the most radical of which 
is arguably the nuclearity attribute, indicating the importance among siblings. 

\begin{figure}[t]
    \centering
    \setlength{\belowcaptionskip}{-10pt}
    \includegraphics[width=.98\linewidth]{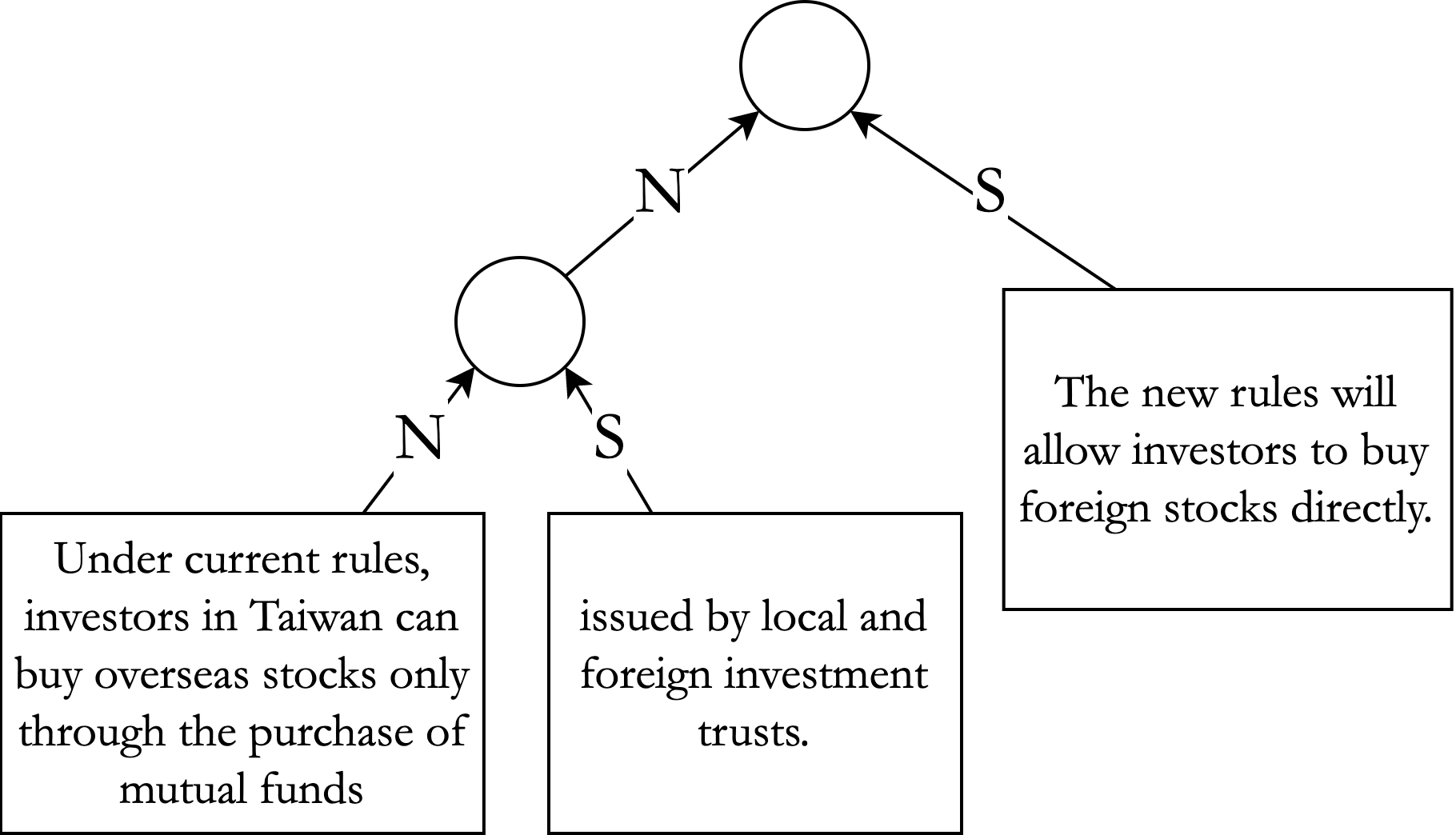}
    \caption{Document $wsj\_0639$ from the RST-DT corpus with inconsistent importance differences between N-S attributions. (The top-level satellite is clearly more central to the overall context than the lower-level satellite. However, both are similarly assigned the satellite attribution by at least one annotator). Top relation: Annotator 1: N-S, Annotator 2: N-N.}
    \label{fig:example_2}
\end{figure}

Intuitively, 
such a coarse (binary) importance assessment does not allow to represent nuanced differences regarding sub-tree importance, which can potentially be critical for downstream tasks. For instance, 
the importance of two nuclei siblings is rather intuitive to interpret. However, having siblings annotated as ``nucleus-satellite" or ``satellite-nucleus" leaves the question on how much more important the nucleus sub-tree is compared to the satellite, as shown in Figure \ref{fig:example_2}. In general, it is unclear (and unlikely) that the actual importance distributions between siblings 
with the same nuclearity attribution 
are consistent.

Based on this observation, we investigate the potential of replacing the binary nuclearity assessment postulated by RST with automatically generated, real-valued importance scores in a new, \textbf{W}eighted-\textbf{RST} framework. In contrast with previous work that has assumed RST and  developed computational models of discourse by simply applying machine learning methods to RST annotated treebanks \cite{ji2014representation, feng2014linear, joty2015codra, li2016discourse, wang2017two, yu2018transition}, we rely on very recent empirical studies showing that weighted ``silver-standard" discourse trees can be inferred 
from auxiliary tasks such as sentiment analysis \cite{huber2020mega} and summarization \cite{xiao-etal-2021-predicting}.

In our 
evaluation, we assess both, computational benefits and linguistic insights. In particular, we find that automatically generated, weighted discourse trees can benefit key NLP downstream tasks. 
We further show that real-valued importance scores (at least partially) align with human annotations and can interestingly also capture uncertainty in human annotators, implying some alignment of the importance distributions with linguistic ambiguity.

\begin{figure*}[t]
    \centering
    \setlength{\belowcaptionskip}{-15pt}
    \includegraphics[width=1\textwidth]{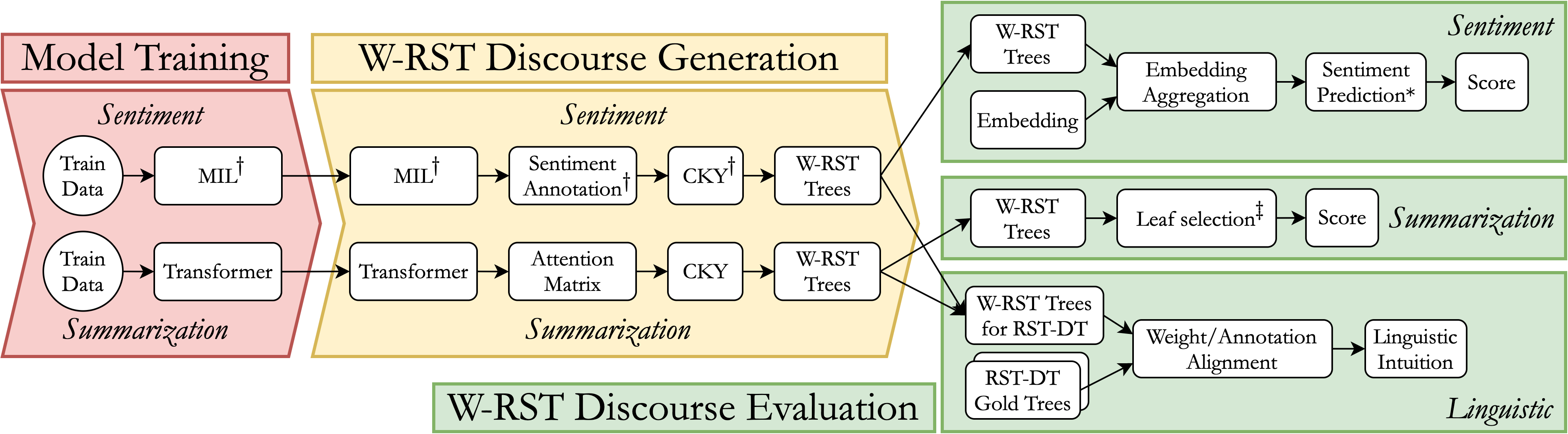}
    \caption{Three phases of our approach to generate weighted RST-style discourse trees. Left and center steps are described in section \ref{wrst_gen}, right component is described in section \ref{eval}. 
    $\dagger$ = As in \citet{huber2020mega}, $\ddagger$ = As in \citet{marcu1999discourse}, $*$ = Sentiment prediction component is a linear combination, mapping the aggregated embedding to the sentiment output. The linear combination has been previously learned on the training portion of the dataset.}
    \label{fig:overall}
\end{figure*}

\section{Related Work}

First introduced by \citet{mann1988rhetorical}, 
the Rhetorical Structure Theory (RST) has been one of the primary guiding theories for discourse analysis \cite{carlson2002rst, subba2009effective, Zeldes2017, gessler2019discourse, liu2019discourse}, discourse parsing \cite{ji2014representation, feng2014linear, joty2015codra, li2016discourse, wang2017two, yu2018transition}, and text planning \cite{torrance2015understanding, gatt2018survey, guz2020towards}. 
The RST framework thereby 
comprehensively describes the organization of a document, guided by the author's communicative goals, 
encompassing three 
components: (1) A projective constituency tree structure, often referred to as the tree span. (2) A nuclearity attribute, assigned to every internal node of the discourse tree, encoding relative importance 
between the nodes' sub-trees, 
with the nucleus 
expressing primary importance and a satellite 
signifying supplementary sub-trees. (3) A relation attribute for every internal node 
describing the relationship between the sub-trees of a node (e.g., Contrast, Evidence, Contradiction). 

Arguably, the weakest aspect of an RST representation is
the nuclearity assessment, which makes a 
too coarse differentiation between primary and secondary importance of sub-trees. 
However, despite its binary assignment of importance and even though the nuclearity attribute is only one of three components of an RST 
tree, it  has major implications for many downstream tasks, as already shown early on by \citet{marcu1999discourse}, using the nuclearity attribute as the key signal in extractive summarization. 
Further work in sentiment analysis \cite{bhatia2015better} also showed the importance of nuclearity for the task 
by first converting the constituency tree into a dependency tree (more aligned with the nuclearity attribute) and then using that tree to predict sentiment more accurately. Both of these results indicate that nuclearity, even in the coarse RST version, 
already contains valuable information. 
Hence, we believe that this coarse-grained classification is reasonable when manually annotating discourse, but see it as a major point of improvement, if a more fine-grained assessment could be correctly assigned. We therefore explore the potential of assigning a weighted nuclearity attribute in this paper.

While plenty of studies have highlighted the important role of discourse for real-world downstream tasks, including summarization, \cite{gerani2014abstractive, xu-etal-2020-discourse, xiao2020we}, sentiment analysis \cite{bhatia2015better,hogenboom2015using,nejat2017exploring} and text classification \cite{ji2017neural}, more critical to our approach is very recent work exploring such connection in the opposite direction. In \citet{huber2020mega}, we exploit sentiment related information to generate ``silver-standard" nuclearity annotated discourse trees, showing their potential on the domain-transfer discourse parsing task. Crucially for our purposes, this approach internally generates real-valued importance-weights for trees.



For the task of extractive summarization, 
we follow our intuition given in \citet{xiao2020we} and \citet{xiao-etal-2021-predicting}, exploiting the connection between summarization and discourse. 
In particular, in \citet{xiao-etal-2021-predicting}, we demonstrate that  the  self-attention matrix learned during the training of a transformer-based summarizer captures valid aspects of constituency and dependency discourse trees. 

To summarize, building on our previous work on creating discourse trees through distant supervision, we take a first step towards generating weighted discourse trees from the sentiment analysis and summarization tasks.

\section{W-RST Treebank Generation}
\label{wrst_gen}
Given the intuition from above, 
we combine information from machine learning approaches with insights 
from linguistics, replacing the human-centered nuclearity assignment with real-valued weights obtained from the 
{\it sentiment analysis} 
and {\it summarization} tasks\footnote{Please note that both tasks use binarized discourse trees, as commonly used in computational models of RST.}. 
An overview of the process 
to generate weighted RST-style discourse trees is shown in Figure \ref{fig:overall}, containing the training phase (left) and the W-RST discourse inference phase (center) described here. The W-RST discourse evaluation 
(right),
is covered in section \ref{eval}.

\begin{figure*}[t]
    \centering
    \includegraphics[width=1\textwidth]{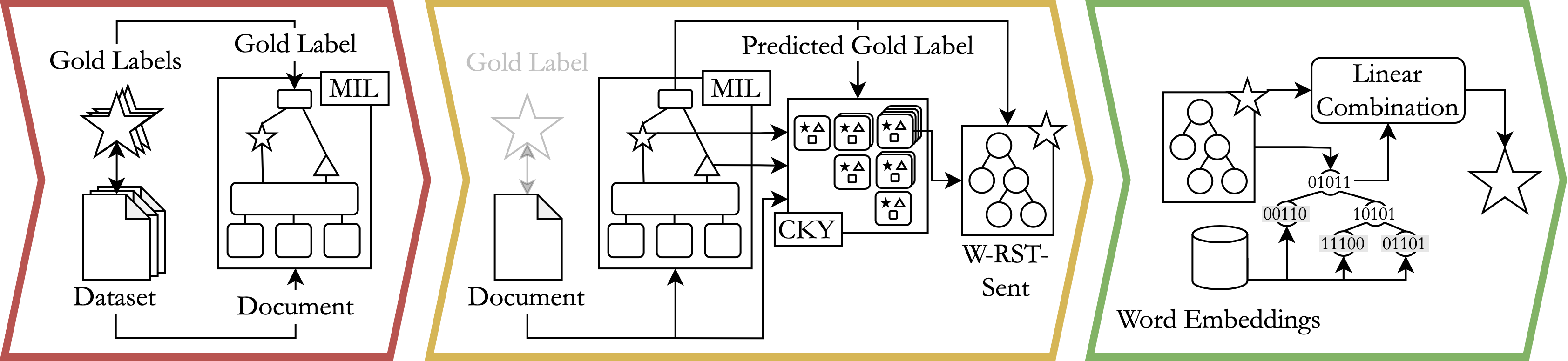}
    \caption{Three phases of our approach.  Left/Center: Detailed view into the generation of weighted RST-style discourse trees using the sentiment analysis downstream task. Right: Sentiment discourse application evaluation}
    \label{fig:sent}
\end{figure*}

\subsection{Weighted Trees from Sentiment 
}
\label{sent_approach}
To generate weighted discourse trees from sentiment, we slightly modify the publicly available code\footnote{Code available at \url{https://github.com/nlpat/MEGA-DT}} presented in \citet{huber2020mega} 
by removing the nuclearity discretization component. 

An overview of our method is shown in  Figure  \ref{fig:overall} (top), while a detailed view 
is presented  in the left and center parts of Figure \ref{fig:sent}. First (on the left), 
we train the Multiple Instance Learning (MIL) model proposed by \citet{angelidis2018multiple} on a corpus with document-level sentiment gold-labels, internally annotating each input-unit (in our case EDUs) with a sentiment- and attention-score.  After the MIL model is trained (center), 
a tuple $(s_i, a_i)$ containing a sentiment score $s_i$ and an attention $a_i$ is extracted for each EDU $i$. Based on these tuples representing leaf nodes, 
the CKY algorithm \cite{jurafsky2014speech} is applied
to find the tree structure to best align with the overall document sentiment, through a bottom-up aggregation approach defined as\footnote{Equations taken from \citet{huber2020mega}}:

\[
s_p = \frac{ s_{l}*a_{l}+s_{r}*a_{r}}{a_{l}+a_{r}}\quad a_p = \frac{a_{l}+a_{r}}{2}
\label{eq:functions}
\]

with nodes $l$ and $r$ as the left and right child-nodes of $p$ respectively. 
The attention scores $(a_l, a_r)$ 
are here interpreted as the importance weights for the respective sub-trees ($w_l = a_l/(a_l+a_r)$ and $w_r = a_r/(a_l+a_r)$), resulting in a complete, normalized and weighted discourse structure as required for W-RST. We call the discourse treebank generated with this approach \textit{W-RST-Sent}.



\subsection{Weighted Trees from Summarization}
\begin{figure*}[t]
    \centering
    \includegraphics[width=1\textwidth]{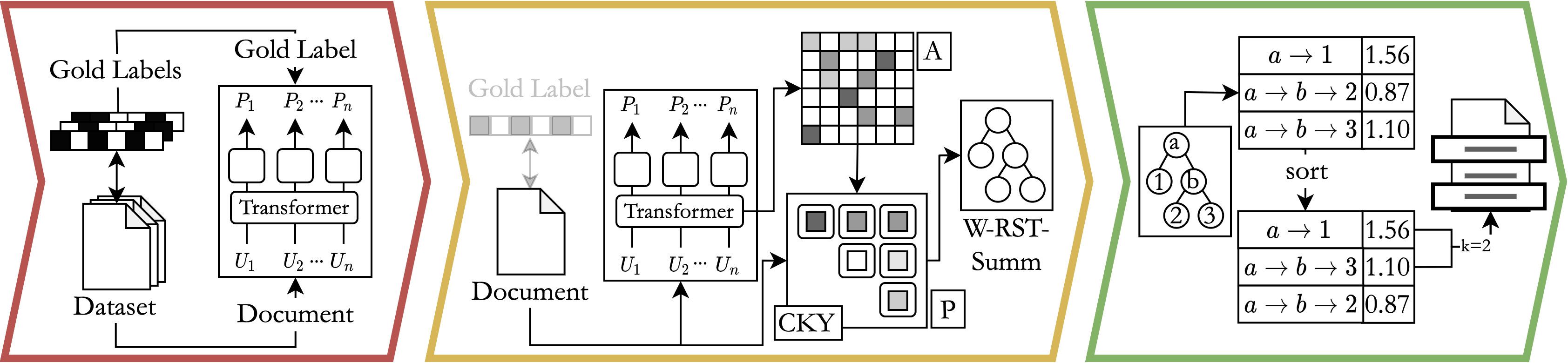}
    \caption{Three phases of our approach.  Left/Center: Detailed view into the generation of weighted RST-style discourse trees using the summarization downstream task. Right: Summarization discourse application evaluation}
    \label{fig:summ}
    \vspace{-3mm}
\end{figure*}


In order to derive weighted discourse trees from a summarization model we follow \citet{xiao-etal-2021-predicting}\footnote{Code available at \url{https://github.com/Wendy-Xiao/summ_guided_disco_parser}},
generating weighted discourse trees from the self-attention matrices of a transformer-based summarization model.
An overview of our method is shown in  Figure  \ref{fig:overall} (bottom), while a detailed view 
is presented in the left and center parts of Figure \ref{fig:summ}. 

We start by training a transformer-based extractive summarization model (left), containing three components: 
(1) A pre-trained BERT EDU Encoder generating EDU embeddings, 
(2) a standard transformer architecture as proposed in \citet{vaswani2017attn} and 
(3) a final classifier, mapping the outputs of the transformer to a probability score for each EDU, indicating whether the EDU should be part of the extractive summary. 

With the trained transformer model, we then extract the self-attention matrix $A$ 
and build a discourse tree in bottom-up fashion (as shown in the center of Figure \ref{fig:summ}). Specifically, 
the self-attention matrix $A$ reflects the relationships between units in the document, where entry $A_{ij}$ measures how much the $i$-th EDU relies on the $j$-th EDU. Given this information, we generate an 
unlabeled constituency tree using the CKY algorithm \cite{jurafsky2014speech}, optimizing the overall tree score, as previously done in \citet{xiao-etal-2021-predicting}.



 
In terms of weight-assignment, given a sub-tree spanning EDUs $i$ to $j$, split into child-constituents at EDU $k$, then 
$\max(A_{i:k, (k+1):j})$, representing the maximal attention value that any EDU in the left constituent is paying to an EDU in the right child-constituent, reflects how much the left sub-tree relies on the right sub-tree, while $\max(A_{(k+1):j, i:k})$ defines how much the right sub-tree depends on the left.
We define the importance-weights of the left ($w_l$) and right ($w_r$) sub-trees as: 
\vspace{-3mm}
\begin{eqnarray*}
w_l &=& 
\max(A_{(k+1):j,i:k})/(w_l+w_r)
\\
w_r &=& 
\max(A_{i:k,(k+1):j})/(w_l+w_r)
\end{eqnarray*}
In this way, the importance scores of the two sub-trees represent a real-valued distribution. In combination with the 
unlabeled structure computation, we generate a weighted discourse tree for each document. We call the discourse treebank generated with the summarization downstream information \textit{W-RST-Summ}.


\section{W-RST Discourse Evaluation}
\label{eval}
To assess the potential of W-RST, we consider two evaluation scenarios
(Figure \ref{fig:overall}, right): (1) Apply 
weighted discourse trees to the tasks of sentiment analysis and summarization and (2) analyze the weight alignment with human annotations.

\subsection{Weight-based Discourse Applications}
\label{info_loss}
In this evaluation scenario, we address the question of whether W-RST trees can support downstream tasks better than traditional RST trees with nuclearity.
Specifically, we leverage the discourse trees learned from sentiment for the sentiment analysis task itself and, similarly, rely on the discourse trees learned from summarization to benefit the summarization task.

 
\subsubsection{Sentiment Analysis}

In order to predict the sentiment of a document in \textit{W-RST-Sent} based on 
its weighted discourse tree, 
we need to introduce an additional source of information to be aggregated according to such tree. Here, we choose word embeddings, as commonly used as an initial transformation in many models tackling the sentiment prediction task \cite{kim2014convolutional,tai2015improved, yang2016hierarchical,adhikari2019rethinking, huber2020sentiment}. To avoid introducing additional confounding factors through sophisticated tree aggregation approaches (e.g. TreeLSTMs \cite{tai2015improved}),
we select a simple method, aiming to directly compare the inferred tree-structures and allowing us to better assess the performance differences originating from the weight/nuclearity attribution (see right step in Figure \ref{fig:sent}). More specifically, we start by computing the average word-embedding for each leaf node ${leaf}_i$ (here containing a single EDU) in the discourse tree. 
\vspace{-3mm}
\[
    {leaf}_i = 
    \sum_{j=0}^{j<|{leaf}_i|}{Emb({word}_{i}^{j})}/|{leaf}_i|
\label{eq:avg}
\]
With $|{leaf}_i|$ as the number of words in leaf $i$, $Emb(\cdot)$ being the embedding lookup and ${word}_{i}^{j}$ representing word $j$ within leaf $i$.
Subsequently, we aggregate constituents, starting from the leaf nodes (with ${leaf}_i$ as embedding constituent $c_i$), according to the weights of the discourse tree. For any two sibling constituents $c_l$ and $c_r$ of the parent sub-tree $c_p$ in the binary tree, we compute

\[
    c_p = c_l*w_l + c_r*w_r
    \label{eq:agg}
\]

with $w_l$ and $w_r$ as the real-valued weight-distribution extracted from the inferred discourse tree and $c_p, c_l$ and $c_r$ as dense encodings. 
We aggregate the complete document in bottom-up fashion, eventually reaching a root node embedding containing a tree-weighted average of the leaf-nodes. 
Given the root-node embedding representing a complete document, a simple Multilayer Perceptron (MLP) trained on the original training portion of the MIL model 
is used to predict the sentiment of the document.

\subsubsection{Summarization}
\label{summ}
In the evaluation step of the summarization model (right of Figure \ref{fig:summ}), we use the weighted discourse tree of a document in \textit{W-RST-Summ} to predict its extractive summary 
by applying an adaptation of the unsupervised summarization method by \citet{marcu1999discourse}. 


We choose this straightforward 
algorithm
over more elaborate and hyper-parameter heavy approaches to avoid confounding factors, since our aim is to evaluate solely the potential of the weighted discourse trees compared to standard RST-style annotations. In the original algorithm, 
a summary is computed based on the nuclearity attribute 
by recursively computing the importance scores for all units as:
\vspace{-3mm}
\[
S_n(u,N)=\begin{cases}
        d_N, & u\in Prom(N)\\
        S(u,C(N))\  s.t.\\
        \quad \ u \in C(N) 
        & otherwise
\end{cases}
\]

where $C(N)$ represents the child of $N$, and $Prom(N)$ is the promotion set of node $N$, which is defined in bottom-up fashion as follows: (1) $Prom$ 
of a leaf node is the leaf node itself. (2) $Prom$ 
of an internal node is the union of the promotion sets of its nucleus children. Furthermore, $d_N$ represents the level of a node $N$, computed as the distance from the level of the lowest leaf-node. This way,
units in the promotion set originating from nodes that are higher up in the discourse tree are amplified in their importance compared to those from lower levels.

As for the \textit{W-RST-Summ} discourse trees with real-valued importance-weights, we adapt Marcu's algorithm by replacing the promotion set with 
real-valued importance scores as shown here:
\vspace{-3mm}
\[
S_{w}(u,N)=\begin{cases}
        d+w_N, & N\ is\ leaf\\
        S_w(u,C(N))+w_N\ ,\\
        \quad \ u \in C(N)
        & otherwise
\end{cases}
\]

Once $S_n$ or $S_{w}$ are computed, 
the top-k units of the highest promotion set or with the highest importance scores respectively are selected into the final summary.


\subsubsection{Nuclearity-attributed Baselines}
\label{baseline}
To test whether the W-RST trees are effectively predicting the downstream tasks, we need to generate traditional RST trees with nuclearity to compare against. 
However, moving from weighted discourse trees to coarse nuclearity 
requires the introduction of a 
threshold. More specifically, while ``nucleus-satellite" and ``satellite-nucleus" assignments can be naturally generated depending on the distinct weights, in order to assign the third ``nucleus-nucleus" class, frequently appearing in RST-style treebanks, we need to specify how close two weights have to be for such configuration to apply. Formally, we set a threshold $t$ as follows:



\begin{center}
\noindent{\bf If:} \quad $|w_l - w_r|<t \quad \rightarrow$ \quad nucleus-nucleus \\
{\bf Else:}\quad{\bf If:}  $w_l > w_r \quad \rightarrow$ \quad nucleus-satellite\\
\textcolor{white}{{\bf Else:}}\quad{\bf If:}  $w_l \leq w_r \quad \rightarrow$ \quad satellite-nucleus
\end{center}


This way, RST-style treebanks with nuclearity attributions can be generated from 
\textit{W-RST-Sent} and \textit{W-RST-Summ} and used for the sentiment analysis and summarization downstream tasks.
For the nuclearity-attributed baseline of the sentiment task, 
we use a similar approach as for the W-RST evaluation procedure, but assign two distinct weights $w_n$ and $w_s$ to the nucleus and satellite child respectively. Since it is not clear how much more important a nucleus node is compared to a satellite using the traditional RST notation, we define the two weights based on the threshold $t$ as:
\vspace{-2mm}
\[
        w_n = 1-(1-2t)/4\;\;\;\;\;\; w_s = (1-2t)/4
        \label{eq:sent_weights}
\]

\begin{figure}[t]
    \centering
    \setlength{\belowcaptionskip}{-10pt}
    \includegraphics[width=1\linewidth]{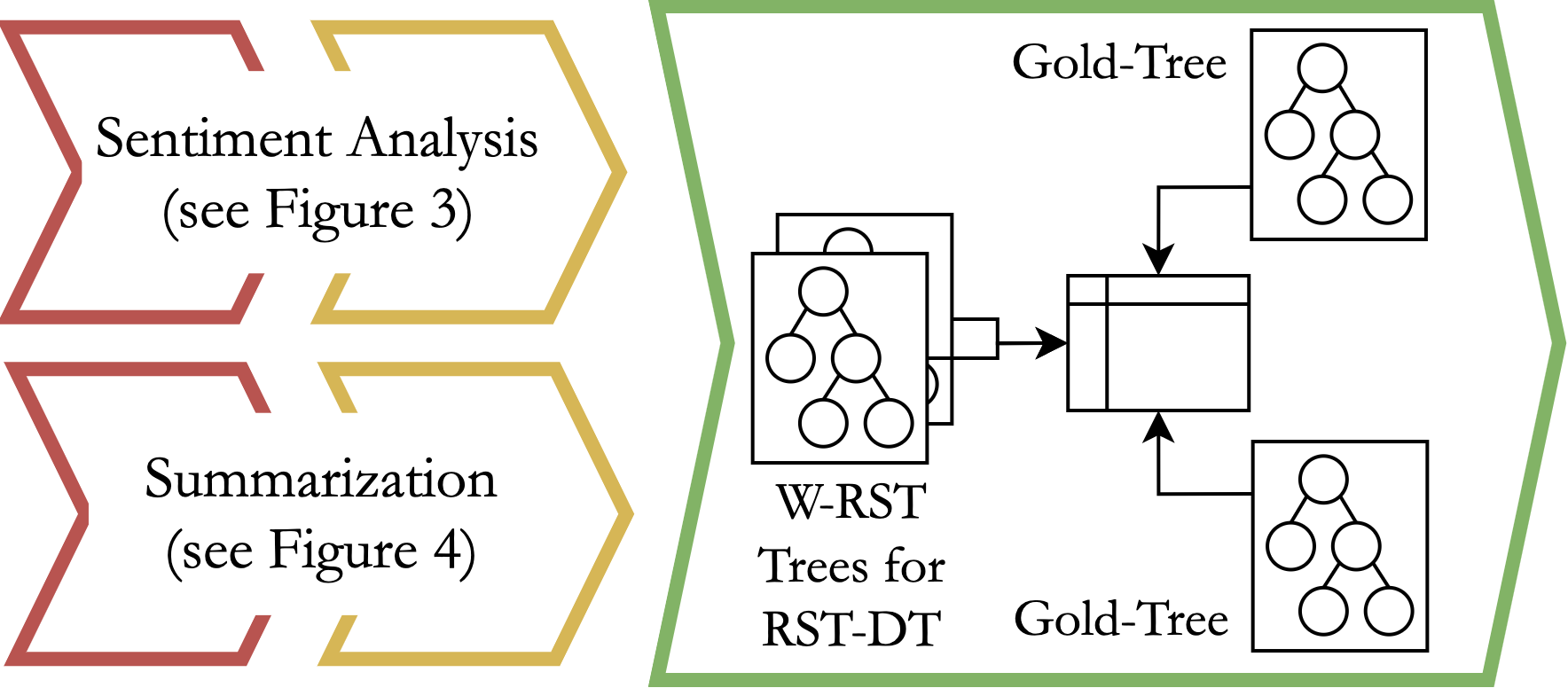}
    \caption{Three phases of our approach.  Left: Generation of \textit{W-RST-Sent/Summ} discourse trees. Right: Linguistic evaluation}
    \label{fig:ling}
\end{figure}

The intuition behind this formulation 
is that for a high threshold $t$ (e.g. $0.8$), the nuclearity needs to be very prominent (the difference between the normalized weights needs to exceed $0.8$), making the nucleus clearly more important than the satellite, while for a small threshold (e.g. $0.1$), even relatively balanced weights (for example $w_l = 0.56, w_r = 0.44$) will be assigned as ``nucleus-satellite", resulting in the potential difference in importance of the siblings to be less eminent.

For the nuclearity-attributed baseline for summarization, we directly apply the original algorithm by \citet{marcu1999discourse} as described in section \ref{summ}. However, 
when using the promotion set to determine which EDUs are added to the summarization, potential ties can occur. Since the discourse tree does not provide any information on how to prioritize those, we randomly select units from the candidates, whenever there is a tie. This avoids exploiting any positional bias in the data (e.g. the lead bias), which would confound the results.


\subsection{Weight Alignment with Human Annotation} 
\label{human_uncertainty}

As for our second W-RST discourse evaluation task, we investigate if the real-valued importance-weights align with human annotations. To be able to explore this scenario, we generate weighted tree annotations for an existing discourse treebank (RST-DT \cite{carlson2002rst}). 
In this evaluation task we verify if: (1) The nucleus in a gold-annotation generally receives more weight than a satellite (i.e. if importance-weights generally favour nuclei over satellites) and, similarly, if nucleus-nucleus relations receive more balanced weights. (2) In accordance with Figure \ref{fig:example_2}, we further explore how well the weights capture the extend to which a relation is dominated by the nucleus. Here, our intuition is that for inconsistent human nuclearity annotations the spread should generally be lower than for consistent annotations, assuming that human misalignment in the discourse annotation indicates ambivalence on the importance of sub-trees.

To test for these two properties, we use discourse documents individually annotated by two human annotators and 
analyze each sub-tree within the doubly-annotated documents with consistent inter-annotator structure assessment for their nuclearity assignment. For each of the 6 possible inter-annotator nuclearity assessments, 
consisting of 3 consistent annotation classes (namely N-N/N-N, N-S/N-S and S-N/S-N) and 3 inconsistent annotation classes (namely N-N/N-S, N-N/S-N and N-S/S-N)\footnote{We don't take the order of annotators into consideration, mapping N-N/N-S and N-S/N-N both onto N-N/N-S.}, we explore the respective weight distribution of the document annotated with the two W-RST tasks -- sentiment analysis and summarization (see Figure \ref{fig:ling}). We compute an average spread $s_{c}$ for each of the 6 inter-annotator nuclearity assessments classes $c$ as:

\vspace{-3mm}
\[
    s_{c} = 
    (\sum_{j=0}^{j<|c|}{w^j_{l}-w^j_{r}})/|c|
\]
With $w^j_{l}$ and $w^j_{r}$ as the weights of the left and right child node of sub-tree $j$ in class $c$, respectively.


\section{Experiments}
\label{experiments}



\subsection{Experimental Setup}
\label{setup}

{\bf Sentiment Analysis:}
We follow our previous approach in 
\citet{huber2020mega} for the model training and W-RST discourse inference steps (left and center in Figure \ref{fig:sent}) using the adapted MILNet model from \citet{angelidis2018multiple} trained with a batch-size of $200$ and $100$ neurons in a single layer bi-directional GRU with $20\%$ dropout for $25$ epochs. 
Next, discourse trees are generated using the best-performing heuristic CKY method with the stochastic exploration-exploitation trade-off from \citet{huber2020mega} (beam size $10$, linear decreasing $\tau$). 
As 
word-embeddings in the W-RST discourse evaluation (right in Figure \ref{fig:sent}), we use GloVe embeddings 
\cite{pennington2014glove}, which previous work \cite{tai2015improved,huber2020sentiment} indicates to be suitable for aggregation in discourse processing.
For training and evaluation of the sentiment analysis task, we use the 5-class Yelp'13 review dataset \cite{tang2015document}.
To compare our approach against the traditional RST approach with nuclearity, we explore the impact of 11 distinct 
thresholds for the baseline described in \S\ref{baseline}, ranging from $0$ to $1$ in $0.1$ intervals. \\
{\bf Summarization:}
To be consistent with RST, our summarizer extracts EDUs instead of sentences from a given document. The model is trained
on the EDU-segmented CNNDM dataset containing EDU-level Oracle labels published by \citet{xu-etal-2020-discourse}.
We further use a pre-trained BERT-base (``uncased") model  
to generate the embeddings of EDUs. The transformer used 
is the standard model with 
6 layers and 8 heads in each layer 
($d=512$). We train the extractive summarizer on the training set of the CNNDM corpus \cite{cnndm} and pick the best attention head using the RST-DT dataset \cite{carlson2002rst} as the development set. 
We test the trees by running the summarization algorithm in \citet{marcu1999discourse} on the test set of the CNNDM dataset, and select the top-6 EDUs based on the importance score to form a summary in natural order. Regarding the baseline model using thresholds, we apply the same 11 thresholds
as for the sentiment analysis 
task.\\

{\bf Weight Alignment with Human Annotation:}
As discussed in \S\ref{human_uncertainty}, this evaluation requires two parallel human generated discourse trees for every document. Luckily, 
in the RST-DT corpus published by \citet{carlson2002rst}, $53$ of the $385$ documents annotated with full RST-style discourse trees are doubly tagged by a second linguist. We use the $53$ documents containing $1,354$ consistent structure annotations between the two analysts to evaluate the linguistic alignment of our generated W-RST documents with human discourse interpretations. Out of the $1,354$ structure-aligned sub-trees, in $1,139$ cases both annotators agreed on the nuclearity attribute, while $215$ times a nuclearity mismatch appeared, as shown in detail in 
Table \ref{tab:stat_human_uncertainty}.

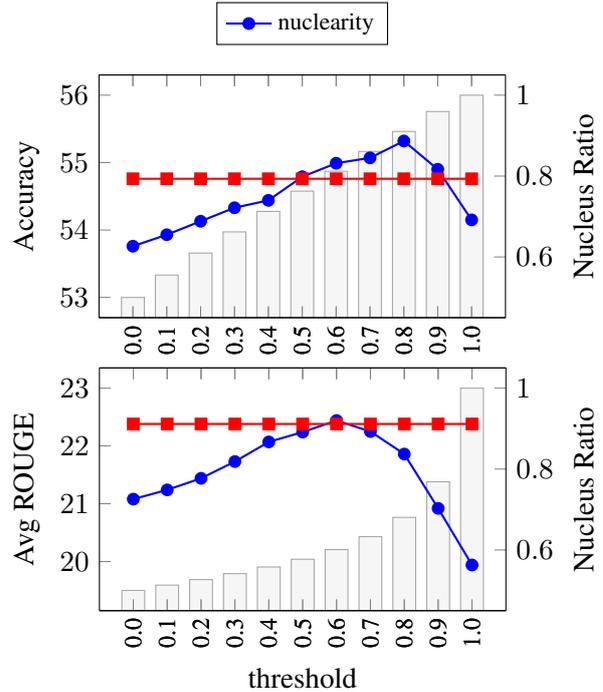
\begin{figure}[t]
    \centering
    \begin{tikzpicture}
        \begin{axis}[
            width=.9\linewidth,
            height=4.8cm,
            ylabel={Nucleus Ratio},
            ylabel near ticks,
            bar width=3mm,
            symbolic x coords={0.0,0.1,0.2,0.3,0.4,0.5,0.6,0.7,0.8,0.9,1.0},
            xtick=data,
            xticklabel style = {font=\small,yshift=0.2ex},
            xticklabel style = {rotate=90,anchor=east},
            axis y line*=right,
        ]
        \addplot[ybar, fill=black!10, opacity=0.35] 
            coordinates {
            (0.0, 0.5)
            (0.1, 0.555334200683038)
            (0.2, 0.609710522036103)
            (0.3, 0.662041795413888)
            (0.4, 0.712678484306391)
            (0.5, 0.762672792323955)
            (0.6, 0.81182306066027)
            (0.7, 0.860953000487884)
            (0.8, 0.910317937876077)
            (0.9, 0.959530004878842)
            (1.0, 1)}; 
        \end{axis}
    
        \begin{axis}[
            axis y line*=left,
            width=.9\linewidth,
            height=4.8cm,
            enlargelimits=0.1,
            ylabel={Accuracy},
            ylabel near ticks,
            ytick={53, 54, 55, 56},
            ymin=53,
            ymax=56,
            symbolic x coords={0.0,0.1,0.2,0.3,0.4,0.5,0.6,0.7,0.8,0.9,1.0},
            xtick=data,
            xticklabel style = {font=\small,yshift=0.2ex},
            xticklabel style = {rotate=90,anchor=east},
            legend style={at={(0.5,1.3)},anchor=north,,legend columns=-1}]
        ]
        \addplot[draw=blue, mark=*, mark options={draw=blue,fill=blue},thick] 
        coordinates {
            (0.0, 53.76)
            (0.1, 53.93)
            (0.2, 54.13)
            (0.3, 54.33)
            (0.4, 54.44)
            (0.5, 54.79)
            (0.6, 54.99)
            (0.7, 55.07)
            (0.8, 55.32)
            (0.9, 54.9)
            (1.0, 54.15)};
            \addlegendentry{\footnotesize{nuclearity}}
    
        \addplot[draw=red, mark=square*, mark options={draw=red,fill=red},thick] 
        coordinates {
            (0.0, 54.76)
            (0.1, 54.76)
            (0.2, 54.76)
            (0.3, 54.76)
            (0.4, 54.76)
            (0.5, 54.76)
            (0.6, 54.76)
            (0.7, 54.76)
            (0.8, 54.76)
            (0.9, 54.76)
            (1.0, 54.76)};
            
        \end{axis}
    \end{tikzpicture}
    
    \begin{tikzpicture}
        \begin{axis}[
            width=.9\linewidth,
            height=4.8cm,
            ylabel={Nucleus Ratio},
            ylabel near ticks,
            bar width=3mm,
            symbolic x coords={0.0,0.1,0.2,0.3,0.4,0.5,0.6,0.7,0.8,0.9,1.0},
            xtick=data,
            xticklabel style = {font=\small,yshift=0.2ex},
            xticklabel style = {rotate=90,anchor=east},
            axis y line*=right,
            xlabel=threshold]
        ]
        \addplot[ybar, fill=black!10, opacity=0.35] 
        coordinates {
            (0.0, 0.5)
            (0.1, 0.513183)
            (0.2, 0.526753)
            (0.3, 0.541541)
            (0.4, 0.557959)
            (0.5, 0.577101)
            (0.6, 0.600890)
            (0.7, 0.632825)
            (0.8, 0.680709)
            (0.9, 0.768473)
            (1.0, 1)};
        \end{axis}
    
        \begin{axis}[
            axis y line*=left,
            width=.9\linewidth,
            height=4.8cm,
            enlargelimits=0.1,
            ylabel={Avg ROUGE},
            ylabel near ticks,
            ymin=19.5,
            ymax=23,
            symbolic x coords={0.0,0.1,0.2,0.3,0.4,0.5,0.6,0.7,0.8,0.9,1.0},
            xtick=data,
            xticklabel style = {font=\small,yshift=0.2ex},
            xticklabel style = {rotate=90,anchor=east},
            legend pos=south west
        ]
        \addplot[draw=blue, mark=*, mark options={draw=blue,fill=blue},thick] 
        coordinates {
            (0.0, 21.08)	
            (0.1, 21.24)	
            (0.2, 21.44)	
            (0.3, 21.73)	
            (0.4, 22.07)	
            (0.5, 22.24)	
            (0.6, 22.44)	
            (0.7, 22.25)	
            (0.8, 21.86)	
            (0.9, 20.92)	
            (1.0, 19.94)};
        
        
    
        \addplot[draw=red, mark=square*, mark options={draw=red,fill=red},thick]  
        coordinates {
            (0.0, 22.38)
            (0.1, 22.38)
            (0.2, 22.38)
            (0.3, 22.38)
            (0.4, 22.38)
            (0.5, 22.38)
            (0.6, 22.38)
            (0.7, 22.38)
            (0.8, 22.38)
            (0.9, 22.38)
            (1.0, 22.38)};
            
        \end{axis}
    \end{tikzpicture}
    \caption{Top: Sentiment Analysis accuracy of the W-RST model compared to the standard RST framework with different thresholds. 
    Bottom: Average ROUGE score (ROUGE-1, -2 and -L) of the W-RST summarization model compared to different thresholds. Full numerical results are shown in Appendix \ref{app:num}.
    }
    \label{fig:results_downstream}
\end{figure}

\begin{table}
    \centering
    \begin{tabular}{|c|c|c|c|}
    \hline
         &  N-N &N-S&S-N\\
    \hline
     N-N   & 273 & 99 & 41 \\
     \hline
     N-S   &-&694 & 75 \\
     \hline
     S-N   &-&-&172 \\
     \hline
    \end{tabular}
    \caption{Statistics on consistently and inconsistently annotated samples of the $1,354$ structure-aligned sub-trees generated by two distinct human annotators.
    }
    \label{tab:stat_human_uncertainty}
\end{table}


\begin{table}[t]
    \centering
    \setlength{\belowcaptionskip}{-17pt}
    \scalebox{0.82}{
    {\renewcommand{\arraystretch}{1.5}
    \begin{tabular}{|M{10mm}|M{10mm}|M{10mm}|M{10mm}|}
        \hline
        {\bf Sent} & N-N & N-S & S-N\\
        \hline
        N-N & 
        \makecell{-0.228\\(106)} &
        \makecell{-0.238\\(33)} &
        \makecell{-0.240\\(19)} \\
        \hline
        N-S 
        & - & 
        \makecell{-0.038\\(325)} & \makecell{-0.044\\(22)}\\
        \hline
        S-N & - & - & 
        \makecell{-0.278\\(115)}\\
        \hline
    \end{tabular}}}
    
    \centering
    \vspace*{0.15 cm}
    \centering
    \scalebox{0.82}{
    {\renewcommand{\arraystretch}{1.5}
    \label{spread_summ_2}
    \begin{tabular}{|M{10mm}|M{10mm}|M{10mm}|M{10mm}|}
        \hline
        {\bf Summ} & N-N & N-S & S-N\\
        \hline
        N-N & 
        \makecell{0.572\\(136)} &
        \makecell{0.604\\(42)} &
        \makecell{0.506\\(25)}\\
        \hline
        N-S 
        & - & 
        \makecell{0.713\\(418)}& 
        \makecell{0.518\\(36)}\\
        \hline
        S-N 
        & - & - & \makecell{0.616\\(134)}\\
        \hline
    \end{tabular}}}
    \caption{Confusion Matrices based on human annotation showing the absolute weight-spread using the Sentiment (top) and Summarization (bottom) tasks on 620  and 791 sub-trees aligned with the human structure prediction, respectively. Cell upper value: Absolute weight spread for the respective combination of human-annotated nuclearities. Lower value (in brackets): Support for this configuration. 
    }
\label{ling_experiment_1}
\end{table}

\begin{table}[t]
    \centering
    \setlength{\belowcaptionskip}{-17pt}
    \scalebox{0.82}{
    {\renewcommand{\arraystretch}{1.5}
    \begin{tabular}{|M{12mm}|M{12mm}|M{12mm}|M{12mm}|}
        \hline
        {\bf Sent} & N-N & N-S & S-N\\
        \hline
        N-N & 
        \cellcolor{myred!36}$\varnothing$-0.36&
        \cellcolor{myred!43}$\varnothing$-0.43&
        \cellcolor{myred!45}$\varnothing$-0.45\\
        \hline
        N-S 
        & - & 
        \cellcolor{mygreen!100}$\varnothing$+1.00&
        \cellcolor{myred!96}$\varnothing$+0.96\\
        \hline
        S-N & - & - & 
        \cellcolor{myred!72}$\varnothing$-0.72\\
        \hline
    \end{tabular}}}
    
    \centering
    \vspace*{0.15 cm}
    \centering
    \scalebox{0.82}{
    {\renewcommand{\arraystretch}{1.5}
    \label{spread_summ}
    \begin{tabular}{|M{12mm}|M{12mm}|M{12mm}|M{12mm}|}
        \hline
        {\bf Summ} & N-N & N-S & S-N\\
        \hline
        N-N & 
        \cellcolor{myred!13}$\varnothing$-0.13&
        \cellcolor{mygreen!13}$\varnothing$+0.13&
        \cellcolor{myred!66}$\varnothing$-0.66\\
        \hline
        N-S 
        & - & 
        \cellcolor{mygreen!100}$\varnothing$+1.00&
        \cellcolor{myred!56}$\varnothing$-0.56\\
        \hline
        S-N 
        & - & - & \cellcolor{mygreen!22}$\varnothing$+0.22\\
        \hline
    \end{tabular}}}
    \caption{Confusion Matrices based on human annotation showing the weight-spread relative to the task-average for Sentiment (top) and Summarization (bottom), aligned with the human structure prediction, respectively. Cell value: Relative weight spread as the divergence from the average spread across all cells in Table \ref{ling_experiment_1}. Color: Positive/Negative divergence, $\varnothing$ = Average value of absolute scores.
    }
\label{ling_experiment_2}
\end{table}
\subsection{Results and Analysis}
\label{results}
The results of the experiments on the \textbf{discourse applications} for sentiment analysis and summarization are shown in Figure \ref{fig:results_downstream}. 
The results 
for sentiment analysis (top) and summarization (bottom) 
thereby show a similar trend: With an increasing threshold and therefore a larger number of 
N-N relations (shown as  grey bars in the Figure), 
the standard RST baseline (blue line) consistently improves for the respective performance measure of both tasks. However, reaching the best performance at a threshold of $0.8$ for sentiment analysis
and $0.6$ for summarization, 
the performance starts to deteriorate.
This general trend seems reasonable, given that N-N relations represent a rather frequent nuclearity connection, however classifying every connection as N-N leads to a severe loss of information. Furthermore, the performance suggests that while the N-N class is important in both cases, the optimal threshold varies depending on the task and potentially also the corpus used, making further task-specific fine-tuning steps mandatory. The weighted discourse trees following our W-RST approach, on the other hand, do not require the definition of a threshold, resulting in a single, promising  performance (red line) for both tasks in Figure \ref{fig:results_downstream}. 
For comparison, we apply the generated trees of a standard RST-style discourse parser (here the Two-Stage parser by \citet{wang2017two}) trained on the RST-DT dataset \cite{carlson2002rst} on both downstream tasks. The fully-supervised parser reaches an accuracy of 44.77\% for sentiment analysis and an average ROUGE score of 26.28 for summarization. While the average ROUGE score of the fully-supervised parser is above the performance of our W-RST results for the summarization task, the accuracy on the sentiment analysis task is well below our approach. We believe that these results are a direct indication of the problematic domain adaptation of fully supervised discourse parsers, where the application on a similar domain (Wall Street Journal articles vs. CNN-Daily Mail articles) leads to superior performances compared to our distantly supervised method, however, with larger domain shifts (Wall Street Journal articles vs. Yelp customer reviews), the performance drops significantly, allowing our distantly supervised model to outperform the supervised discourse trees for the downstream task.
Arguably, this indicates that although our weighted approach is still not competitive with fully-supervised models in the same domain, it is the most promising solution available for cross-domain discourse parsing.

With respect to exploring the \textbf{weight alignment with human annotations}, we show a set of confusion matrices based on human annotation for each W-RST discourse generation task on the absolute and relative weight-spread in Tables~\ref{ling_experiment_1} and \ref{ling_experiment_2} respectively. The results for the sentiment analysis task are shown on the top of both tables, while the performance for the summarization task is shown at the bottom. 
For instance, the top right cell of the upper confusion matrix in  Table~\ref{ling_experiment_1} shows that for $19$ sub-trees in the doubly annotated subset of RST-DT 
one of the annotators labelled the sub-tree with a nucleus-nucleus nuclearity attribution, while the second annotator identified it as satellite-nucleus. The average weight spread (see \S\ref{human_uncertainty}) 
for those $19$ sub-trees is $-0.24$. 
Regarding Table~\ref{ling_experiment_2}, we subtract the average spread across Table~\ref{ling_experiment_1} defined as ${\varnothing} = \sum_{c_i \in C}{(c_i)}/|C|$ (with $C=\{c_1, c_2, ... c_6\}$ containing the cell values in the upper triangle matrix) from each cell value $c_i$ and normalize by ${max} = max_{c_i \in C}(|c_i-\varnothing|)$, with $\varnothing = -0.177$ and $max = 0.1396$ across the top table. Accordingly, we transform the $-0.24$ in the top right cell into $(-0.24 - avg)/max = -0.45$.

Moving to the analysis of the results, we find the following trends in this experiment: 
(1) As presented in Table~\ref{ling_experiment_1}, the sentiment analysis task tends to strongly over-predict S-N (i.e., $w_{l} << w_{r}$), leading to negative spreads in all cells. In contrast, the summarization task is heavily skewed towards N-S assignments (i.e., $w_{l} >> w_{r}$), leading to exclusively positive spreads. We believe both trends are consistent with the intrinsic properties of the tasks, given that the general structure of reviews tends to become more important towards the end of a review (leading to increased S-N assignments), while for summarization, the lead bias potentially produces the overall strong nucleus-satellite trend. 
(2) To investigate the relative weight spreads for different human annotations (i.e., between cells) beyond the trends shown in Table~\ref{ling_experiment_1}, 
we normalize values within a table by subtracting the average and scaling between $[-1, 1]$. As a result, Table~\ref{ling_experiment_2} shows the relative weight spread for different human annotations. Apart from the general trends described in Table~\ref{ling_experiment_1}, the consistently annotated samples of the two linguists (along the diagonal of the confusion matrices) align reasonably. The most positive weight spread is consistently found in the agreed-upon nucleus-satellite case, while the nucleus-nucleus annotation has, as expected, the lowest divergence (i.e., closest to zero) along the diagonal in Table~\ref{ling_experiment_2}. 
(3) Regarding the inconsistently annotated samples (shown in the triangle matrix above the diagonal) it becomes clear that in the sentiment analysis model the values for the N-N/N-S and N-N/S-N annotated samples (top row in Table~\ref{ling_experiment_2}) are relatively close to the average value. This indicates that, similar to the nucleus-nucleus case, the weights are also ambivalent, with 
the N-N/N-S value (top center) slightly larger than the value for N-N/S-N (top right).
The N-S/S-N case for the sentiment analysis model is less aligned with our intuition, showing a strongly negative weight-spread (i.e. $w_{l} << w_{r}$) where we would have expected a more ambivalent result with $w_{l} \approx w_{r}$ (however, aligned with the overall trend shown in Table~\ref{ling_experiment_1}). For summarization, we see a very similar trend with the values for N-N/N-S and N-N/S-N annotated samples. Again, both values are close to the average, with the N-N/N-S cell showing a more positive spread than N-N/S-N. However for summarization, the consistent satellite-nucleus annotation (bottom right cell) seems misaligned with the rest of the table, following instead the general trend for summarization described in Table~\ref{ling_experiment_1}. All in all, the results suggest that the values in most cells are well aligned with what we would expect regarding the relative spread. Interestingly, human uncertainty appears to be reasonably captured in the weights, which seem to contain more fine grained information about the relative importance of sibling sub-trees.
\section{Conclusion and Future Work}
We propose W-RST as a new discourse framework, where the binary nuclearity assessment postulated by RST is replaced with more expressive weights, that can be automatically generated from auxiliary tasks. A series of experiments indicate that W-RST is beneficial to the two key NLP downstream tasks of sentiment analysis and summarization. Further, we show that W-RST trees 
interestingly align with the uncertainty of human annotations.

For the future, we plan to develop a neural discourse parser that learns to predict importance weights instead of nuclearity attributions when trained on large W-RST treebanks. More longer term, we want to explore other aspects of RST that can be refined in light of empirical results, plan to integrate our results into state-of-the-art sentiment analysis and summarization approaches (e.g. \citet{xu-etal-2020-discourse}) and generate parallel W-RST structures in a multi-task manner to improve the generality of the discourse trees. 

\section*{Acknowledgments}
We thank the anonymous reviewers for their insightful comments. This research was supported by the Language \& Speech Innovation Lab of Cloud BU, Huawei Technologies Co., Ltd and the Natural Sciences and Engineering Research Council of Canada (NSERC). Nous remercions le Conseil de recherches en sciences naturelles et en génie du Canada (CRSNG) de son soutien.

\bibliographystyle{acl_natbib}
\bibliography{acl2021}

\begin{thebibliography}{37}
\expandafter\ifx\csname natexlab\endcsname\relax\def\natexlab#1{#1}\fi

\bibitem[{Adhikari et~al.(2019)Adhikari, Ram, Tang, and
  Lin}]{adhikari2019rethinking}
Ashutosh Adhikari, Achyudh Ram, Raphael Tang, and Jimmy Lin. 2019.
\newblock Rethinking complex neural network architectures for document
  classification.
\newblock In \emph{Proceedings of the 2019 Conference of the North American
  Chapter of the Association for Computational Linguistics: Human Language
  Technologies, Volume 1 (Long and Short Papers)}, pages 4046--4051.

\bibitem[{Angelidis and Lapata(2018)}]{angelidis2018multiple}
Stefanos Angelidis and Mirella Lapata. 2018.
\newblock Multiple instance learning networks for fine-grained sentiment
  analysis.
\newblock \emph{Transactions of the Association for Computational Linguistics},
  6:17--31.

\bibitem[{Bhatia et~al.(2015)Bhatia, Ji, and Eisenstein}]{bhatia2015better}
Parminder Bhatia, Yangfeng Ji, and Jacob Eisenstein. 2015.
\newblock Better document-level sentiment analysis from {RST} discourse
  parsing.
\newblock In \emph{Proceedings of the 2015 Conference on Empirical Methods in
  Natural Language Processing}, pages 2212--2218.

\bibitem[{Carlson et~al.(2002)Carlson, Okurowski, and Marcu}]{carlson2002rst}
Lynn Carlson, Mary~Ellen Okurowski, and Daniel Marcu. 2002.
\newblock \emph{{RST} discourse treebank}.
\newblock Linguistic Data Consortium, University of Pennsylvania.

\bibitem[{Feng and Hirst(2014)}]{feng2014linear}
Vanessa~Wei Feng and Graeme Hirst. 2014.
\newblock A linear-time bottom-up discourse parser with constraints and
  post-editing.
\newblock In \emph{Proceedings of the 52nd Annual Meeting of the Association
  for Computational Linguistics (Volume 1: Long Papers)}, pages 511--521.

\bibitem[{Gatt and Krahmer(2018)}]{gatt2018survey}
Albert Gatt and Emiel Krahmer. 2018.
\newblock Survey of the state of the art in natural language generation: Core
  tasks, applications and evaluation.
\newblock \emph{Journal of Artificial Intelligence Research}, 61:65--170.

\bibitem[{Gerani et~al.(2014)Gerani, Mehdad, Carenini, Ng, and
  Nejat}]{gerani2014abstractive}
Shima Gerani, Yashar Mehdad, Giuseppe Carenini, Raymond~T Ng, and Bita Nejat.
  2014.
\newblock Abstractive summarization of product reviews using discourse
  structure.
\newblock In \emph{Proceedings of the 2014 conference on empirical methods in
  natural language processing (EMNLP)}, pages 1602--1613.

\bibitem[{Gessler et~al.(2019)Gessler, Liu, and Zeldes}]{gessler2019discourse}
Luke Gessler, Yang~Janet Liu, and Amir Zeldes. 2019.
\newblock A discourse signal annotation system for rst trees.
\newblock In \emph{Proceedings of the Workshop on Discourse Relation Parsing
  and Treebanking 2019}, pages 56--61.

\bibitem[{Guz and Carenini(2020)}]{guz2020towards}
Grigorii Guz and Giuseppe Carenini. 2020.
\newblock Towards domain-independent text structuring trainable on large
  discourse treebanks.
\newblock In \emph{Proceedings of the 2020 Conference on Empirical Methods in
  Natural Language Processing: Findings}, pages 3141--3152.

\bibitem[{Hogenboom et~al.(2015)Hogenboom, Frasincar, De~Jong, and
  Kaymak}]{hogenboom2015using}
Alexander Hogenboom, Flavius Frasincar, Franciska De~Jong, and Uzay Kaymak.
  2015.
\newblock Using rhetorical structure in sentiment analysis.
\newblock \emph{Commun. ACM}, 58(7):69--77.

\bibitem[{Huber and Carenini(2020{\natexlab{a}})}]{huber2020sentiment}
Patrick Huber and Giuseppe Carenini. 2020{\natexlab{a}}.
\newblock From sentiment annotations to sentiment prediction through discourse
  augmentation.
\newblock In \emph{Proceedings of the 28th International Conference on
  Computational Linguistics}, pages 185--197.

\bibitem[{Huber and Carenini(2020{\natexlab{b}})}]{huber2020mega}
Patrick Huber and Giuseppe Carenini. 2020{\natexlab{b}}.
\newblock {MEGA RST} discourse treebanks with structure and nuclearity from
  scalable distant sentiment supervision.
\newblock In \emph{Proceedings of the 2020 Conference on Empirical Methods in
  Natural Language Processing (EMNLP)}, pages 7442--7457.

\bibitem[{Ji and Eisenstein(2014)}]{ji2014representation}
Yangfeng Ji and Jacob Eisenstein. 2014.
\newblock Representation learning for text-level discourse parsing.
\newblock In \emph{Proceedings of the 52nd Annual Meeting of the Association
  for Computational Linguistics (Volume 1: Long Papers)}, volume~1, pages
  13--24.

\bibitem[{Ji and Smith(2017)}]{ji2017neural}
Yangfeng Ji and Noah~A Smith. 2017.
\newblock Neural discourse structure for text categorization.
\newblock In \emph{Proceedings of the 55th Annual Meeting of the Association
  for Computational Linguistics (Volume 1: Long Papers)}, pages 996--1005.

\bibitem[{Joty et~al.(2015)Joty, Carenini, and Ng}]{joty2015codra}
Shafiq Joty, Giuseppe Carenini, and Raymond~T Ng. 2015.
\newblock {CODRA}: A novel discriminative framework for rhetorical analysis.
\newblock \emph{Computational Linguistics}, 41(3).

\bibitem[{Jurafsky and Martin(2014)}]{jurafsky2014speech}
Dan Jurafsky and James~H Martin. 2014.
\newblock \emph{Speech and language processing}, volume~3.
\newblock Pearson London.

\bibitem[{Kim(2014)}]{kim2014convolutional}
Yoon Kim. 2014.
\newblock Convolutional neural networks for sentence classification.
\newblock In \emph{Proceedings of the 2014 Conference on Empirical Methods in
  Natural Language Processing (EMNLP)}, pages 1746--1751.

\bibitem[{Li et~al.(2016)Li, Li, and Chang}]{li2016discourse}
Qi~Li, Tianshi Li, and Baobao Chang. 2016.
\newblock Discourse parsing with attention-based hierarchical neural networks.
\newblock In \emph{Proceedings of the 2016 Conference on Empirical Methods in
  Natural Language Processing}, pages 362--371.

\bibitem[{Liu and Zeldes(2019)}]{liu2019discourse}
Yang Liu and Amir Zeldes. 2019.
\newblock Discourse relations and signaling information: Anchoring discourse
  signals in rst-dt.
\newblock \emph{Proceedings of the Society for Computation in Linguistics},
  2(1):314--317.

\bibitem[{Mann and Thompson(1988)}]{mann1988rhetorical}
William~C Mann and Sandra~A Thompson. 1988.
\newblock Rhetorical structure theory: Toward a functional theory of text
  organization.
\newblock \emph{Text}, 8(3):243--281.

\bibitem[{Marcu(1999)}]{marcu1999discourse}
Daniel Marcu. 1999.
\newblock Discourse trees are good indicators of importance in text.
\newblock \emph{Advances in automatic text summarization}, 293:123--136.

\bibitem[{Nallapati et~al.(2016)Nallapati, Zhou, dos Santos,
  GuÌ‡l{\c{c}}ehre, and Xiang}]{cnndm}
Ramesh Nallapati, Bowen Zhou, Cicero dos Santos, {\c{C}}a{\u{g}}lar
  GuÌ‡l{\c{c}}ehre, and Bing Xiang. 2016.
\newblock Abstractive text summarization using sequence-to-sequence {RNN}s and
  beyond.
\newblock In \emph{Proceedings of The 20th {SIGNLL} Conference on Computational
  Natural Language Learning}, pages 280--290. Association for Computational
  Linguistics.

\bibitem[{Nejat et~al.(2017)Nejat, Carenini, and Ng}]{nejat2017exploring}
Bita Nejat, Giuseppe Carenini, and Raymond Ng. 2017.
\newblock Exploring joint neural model for sentence level discourse parsing and
  sentiment analysis.
\newblock In \emph{Proceedings of the 18th Annual SIGdial Meeting on Discourse
  and Dialogue}, pages 289--298.

\bibitem[{Pennington et~al.(2014)Pennington, Socher, and
  Manning}]{pennington2014glove}
Jeffrey Pennington, Richard Socher, and Christopher~D. Manning. 2014.
\newblock Glove: Global vectors for word representation.
\newblock In \emph{Empirical Methods in Natural Language Processing (EMNLP)},
  pages 1532--1543.

\bibitem[{Prasad et~al.(2008)Prasad, Dinesh, Lee, Miltsakaki, Robaldo, Joshi,
  and Webber}]{prasadpenn}
Rashmi Prasad, Nikhil Dinesh, Alan Lee, Eleni Miltsakaki, Livio Robaldo,
  Aravind Joshi, and Bonnie Webber. 2008.
\newblock The penn discourse treebank 2.0.
\newblock \emph{LREC}.

\bibitem[{Subba and Di~Eugenio(2009)}]{subba2009effective}
Rajen Subba and Barbara Di~Eugenio. 2009.
\newblock An effective discourse parser that uses rich linguistic information.
\newblock In \emph{Proceedings of Human Language Technologies: The 2009 Annual
  Conference of the North American Chapter of the Association for Computational
  Linguistics}, pages 566--574. Association for Computational Linguistics.

\bibitem[{Tai et~al.(2015)Tai, Socher, and Manning}]{tai2015improved}
Kai~Sheng Tai, Richard Socher, and Christopher~D Manning. 2015.
\newblock Improved semantic representations from tree-structured long
  short-term memory networks.
\newblock In \emph{Proceedings of the 53rd Annual Meeting of the Association
  for Computational Linguistics and the 7th International Joint Conference on
  Natural Language Processing (Volume 1: Long Papers)}, pages 1556--1566.

\bibitem[{Tang et~al.(2015)Tang, Qin, and Liu}]{tang2015document}
Duyu Tang, Bing Qin, and Ting Liu. 2015.
\newblock Document modeling with gated recurrent neural network for sentiment
  classification.
\newblock In \emph{Proceedings of the 2015 conference on empirical methods in
  natural language processing}, pages 1422--1432.

\bibitem[{Torrance(2015)}]{torrance2015understanding}
Mark Torrance. 2015.
\newblock Understanding planning in text production.
\newblock \emph{Handbook of writing research}, pages 1682--1690.

\bibitem[{Vaswani et~al.(2017)Vaswani, Shazeer, Parmar, Uszkoreit, Jones,
  Gomez, Kaiser, and Polosukhin}]{vaswani2017attn}
Ashish Vaswani, Noam Shazeer, Niki Parmar, Jakob Uszkoreit, Llion Jones,
  Aidan~N Gomez, {\L}ukasz Kaiser, and Illia Polosukhin. 2017.
\newblock Attention is all you need.
\newblock In \emph{Proceedings of the 31st International Conference on Neural
  Information Processing Systems}, pages 6000--6010.

\bibitem[{Wang et~al.(2017)Wang, Li, and Wang}]{wang2017two}
Yizhong Wang, Sujian Li, and Houfeng Wang. 2017.
\newblock A two-stage parsing method for text-level discourse analysis.
\newblock In \emph{Proceedings of the 55th Annual Meeting of the Association
  for Computational Linguistics (Volume 2: Short Papers)}, pages 184--188.

\bibitem[{Xiao et~al.(2020)Xiao, Huber, and Carenini}]{xiao2020we}
Wen Xiao, Patrick Huber, and Giuseppe Carenini. 2020.
\newblock Do we really need that many parameters in transformer for extractive
  summarization? discourse can help!
\newblock In \emph{Proceedings of the First Workshop on Computational
  Approaches to Discourse}, pages 124--134.

\bibitem[{Xiao et~al.(2021)Xiao, Huber, and
  Carenini}]{xiao-etal-2021-predicting}
Wen Xiao, Patrick Huber, and Giuseppe Carenini. 2021.
\newblock Predicting discourse trees from transformer-based neural summarizers.
\newblock In \emph{Proceedings of the 2021 Conference of the North American
  Chapter of the Association for Computational Linguistics: Human Language
  Technologies}, pages 4139--4152, Online. Association for Computational
  Linguistics.

\bibitem[{Xu et~al.(2020)Xu, Gan, Cheng, and Liu}]{xu-etal-2020-discourse}
Jiacheng Xu, Zhe Gan, Yu~Cheng, and Jingjing Liu. 2020.
\newblock Discourse-aware neural extractive text summarization.
\newblock In \emph{Proceedings of the 58th Annual Meeting of the Association
  for Computational Linguistics}, pages 5021--5031. Association for
  Computational Linguistics.

\bibitem[{Yang et~al.(2016)Yang, Yang, Dyer, He, Smola, and
  Hovy}]{yang2016hierarchical}
Zichao Yang, Diyi Yang, Chris Dyer, Xiaodong He, Alex Smola, and Eduard Hovy.
  2016.
\newblock Hierarchical attention networks for document classification.
\newblock In \emph{Proceedings of the 2016 conference of the North American
  chapter of the association for computational linguistics: human language
  technologies}, pages 1480--1489.

\bibitem[{Yu et~al.(2018)Yu, Zhang, and Fu}]{yu2018transition}
Nan Yu, Meishan Zhang, and Guohong Fu. 2018.
\newblock Transition-based neural rst parsing with implicit syntax features.
\newblock In \emph{Proceedings of the 27th International Conference on
  Computational Linguistics}, pages 559--570.

\bibitem[{Zeldes(2017)}]{Zeldes2017}
Amir Zeldes. 2017.
\newblock The {GUM} corpus: Creating multilayer resources in the classroom.
\newblock \emph{Language Resources and Evaluation}, 51(3):581--612.

\end{thebibliography}

\appendix
\section{Numeric Results}
\label{app:num}
The numeric results of our W-RST approach for the sentiment analysis and summarization downstream tasks presented in Figure \ref{fig:results_downstream} are shown in Table~\ref{tab:numerical_results} below, along with the threshold-based approach, as well as the supervised parser.
\begin{table}[h!]
    \centering
    \resizebox{\linewidth}{!}{
    \begin{tabular}{|c|r|r r r|}
    \hline
        \multirow{2}{*}{Approach} & \multicolumn{1}{c|}{Sentiment} & \multicolumn{3}{c|}{Summarization}\\
         & Accuracy & R-1 & R-2 & R-L \\
        \hline\hline
        \multicolumn{5}{|c|}{Nuclearity with Threshold}\\
        \hline\hline
        t = 0.0 & 53.76 & 28.22 & 8.58 & 26.45 \\
        t = 0.1 & 53.93 & 28.41 & 8.69 & 26.61 \\
        t = 0.2 & 54.13 & 28.64 & 8.85 & 26.83 \\
        t = 0.3 & 54.33 & 28.96 & 9.08 & 27.14 \\
        t = 0.4 & 54.44 & 29.36 & 9.34 & 27.51 \\
        t = 0.5 & 54.79 & 29.55 & 9.50 & 27.68 \\
        t = 0.6 & 54.99 & 29.78 & 9.65 & 27.90 \\
        t = 0.7 & 55.07 & 29.57 & 9.45 & 27.74 \\
        t = 0.8 & 55.32 & 29.18 & 9.08 & 27.32 \\
        t = 0.9 & 54.90 & 28.11 & 8.29 & 26.35 \\
        t = 1.0 & 54.15 & 26.94 & 7.60 & 25.27 \\
        \hline\hline
        \multicolumn{5}{|c|}{Our Weighted RST Framework}\\
        \hline\hline
        weighted & 54.76 & 29.70 & 9.58 & 27.85 \\
        \hline\hline
        \multicolumn{5}{|c|}{Supervised Training on RST-DT}\\
        \hline\hline
        supervised & 44.77 & 34.20 & 12.77 & 32.09\\
        \hline
    \end{tabular}
    }
    \caption{Results of the W-RST approach compared to threshold-based nuclearity assignments and supervised training on RST-DT.}
    \label{tab:numerical_results}
\end{table}

\end{document}